\documentclass{llncs}
\usepackage{array,multirow,graphicx}
\usepackage{balance}
\usepackage[english]{babel}
\usepackage{lipsum}
\usepackage{enumitem}
\usepackage{xfrac}
\usepackage{cite}
\usepackage{amsmath,amssymb,amsfonts}
\usepackage{algorithmic}
\usepackage{graphicx}
\usepackage{textcomp}
\def\BibTeX{{\rm B\kern-.05em{\sc i\kern-.025em b}\kern-.08em
    T\kern-.1667em\lower.7ex\hbox{E}\kern-.125emX}}
\usepackage[utf8]{inputenc}
\usepackage{csquotes}
\usepackage{url}
\usepackage{stfloats}
\usepackage{cases}
\usepackage{array}
 \usepackage{pifont}
\usepackage[para]{footmisc}
\newcolumntype{P}[1]{>{\centering\arraybackslash}p{#1}}
\usepackage{graphicx}
\def\infinity{\rotatebox{90}{8}}
\usepackage{algorithmic}
\usepackage{subcaption}

\usepackage{url}
\usepackage{lineno,hyperref}

\hypersetup{
    colorlinks=true,
    linkcolor=blue,
    urlcolor=blue,
		citecolor=blue,
}

\begin{document}

\title{A Regression Tsetlin Machine with\\Integer Weighted Clauses for\\Compact Pattern Representation}

\author{K. Darshana Abeyrathna \and
Ole-Christoffer Granmo \and Morten~Goodwin}
\institute{Centre for Artificial Intelligence Research, University of Agder, Grimstad, Norway
\email{\{darshana.abeyrathna, ole.granmo, morten.goodwin\}}@uia.no}
\maketitle 

\begin{abstract}
The Regression Tsetlin Machine (RTM) addresses the lack of interpretability impeding state-of-the-art nonlinear regression models. It does this by using conjunctive clauses in propositional logic to capture the underlying non-linear frequent patterns in the data. These, in turn, are combined into a continuous output through summation, akin to a linear regression function, however, with non-linear components and unity weights. Although the RTM has solved non-linear regression problems with competitive accuracy, the resolution of the output is proportional to the number of clauses employed. This means that computation cost increases with resolution. To reduce this problem, we here introduce integer weighted RTM clauses. Our integer weighted clause is a compact representation of multiple clauses that capture the same sub-pattern --- $N$ repeating clauses are turned into one, with an integer weight $N$. This reduces computation cost $N$ times, and increases interpretability through a sparser representation. We further introduce a novel learning scheme that allows us to simultaneously learn both the clauses and their weights, taking advantage of so-called stochastic searching on the line. We evaluate the potential of the integer weighted RTM empirically using six artificial datasets. The results show that the integer weighted RTM is able to acquire on par or better accuracy using significantly less computational resources compared to regular RTMs. We further show that integer weights yield improved accuracy over real-valued ones.
\end{abstract}


\section{Introduction}
The recently introduced Regression Tsetlin Machine (RTM) \cite{abeyrathnartm, abeyrathnartm2} is a propositional logic based approach to interpretable non-linear regression, founded on the Tsetlin Machine (TM) \cite{Ole1}. Being based on disjunctive normal form, like Karnaugh maps, the TM can map an exponential number of input feature value combinations to an appropriate output \cite{phoulady2020weighted}. Recent research reports several distinct properties: i) the conjunctive clauses that the TM produces have an interpretable form, similar to the branches in a decision tree (e.g., \textbf{if} X \textbf{satisfies} condition A \textbf{and not} condition B \textbf{then} Y = 1) \cite{berge2019}. ii) For small-scale pattern recognition problems where the complete TM logic maps to a single circuit, energy consumption is up to three orders of magnitude lower than corresponding neural network architectures, and inference speed is up to four orders of magnitude faster \cite{wheeldon2020hardware}; iii) Like neural networks, the TM can be used in convolution, providing competitive memory usage, computation speed, and accuracy results on MNIST, F-MNIST and K-MNIST, in comparison with simple 4-layer CNNs, K-Nereast Neighbors, SVMs, Random Forests, Gradient Boosting,  BinaryConnect, Logistic Circuits, ResNet, and a recent FPGA-accelerated Binary CNN  \cite{granmo2019convolutional}. Lately, Phoulady et al. have improved the TM computation- and accuracy-wise by introducing real-valued weighted clauses \cite{phoulady2020weighted}. Gorji et al. have simplified hyper-parameter search by means of multi-granular clauses, eliminating the specificity parameter \cite{gorji2019multigranular}.

\textbf{Paper contributions:} While the RTM compares favourably with Regression Trees, Random Forests and Support Vector Regression \cite{abeyrathnartm2}, regression resolution is proportional to the number of conjunctive clauses employed. In other words, computation cost and memory usage grows proportionally with resolution. Building upon the Weighted TM (WTM) by Phoulady et al., \cite{phoulady2020weighted}, this paper introduces weights to the RTM scheme. However, while the WTM uses real-valued weights for classification, we instead propose a novel scheme based on \emph{integer} weights, targeting \emph{regression}. In brief, we use the stochastic searching on the line approach pioneered by Oommen in 1997 \cite{oommen1997stochastic} to eliminate multiplication from the weight updating, relying purely on increment and decrement operations. In addition to the computational benefits this entails, we also argue that integer weighted clauses are more interpretable than real-valued ones because they can be seen as multiple copies of the same clause. Finally, our scheme does not introduce additional hyper-parameters, whereas the WTM relies on  weight learning speed. Empirically, our proposed scheme aids the RTM to achieve similar or better accuracy with significantly fewer clauses, while further enhancing the interpretability of the RTM.

{\bf Paper Organization:} The remainder of the paper is organized as follows. In Section 2, the basics of RTMs are provided. Then, in Section 3, the SPL problem and its solution are explained. The main contribution of this paper, the integer weighting scheme for the RTM, is presented in detail in Section 4 and evaluated empirically using six different artificial datasets in Section 5. We conclude our work in Section 6.

\section{The Regression Tsetlin Machine (RTM)}
The RTM performs regression based on formulas in propositional logic. In all brevity, the input to an RTM is a vector $\mathbf{X}$ of $o$ propositional variables $x_k$, $\mathbf{X} \in \{0,1\}^{o}$. These are further augmented with their negated counterparts $\bar x_k = 1 - x_k$ to form a vector of literals:
$\mathbf{L} = [x_1, \ldots , x_o, \bar x_1, \ldots , \bar x_o] = [l_1, \ldots , l_{2o}]$. In contrast to a regular TM, the output of an RTM is real-valued, normalized to the domain $y \in [0, 1]$.

{\bf Regression Function:}  The regression function of an RTM is simply a linear summation of products, where the products are built from the literals:
\begin{equation}\label{Eq0}
y = \frac{1}{T} \sum_{j=1}^m \prod_{k \in I_j} l_k. 
\end{equation}
Above, the index $j$ refers to one particular product of literals, defined by the subset $I_j$ of literal indexes. If we e.g. have two propositional variables $x_1$  and $x_2$, the literal index sets $I_1 = \{1, 4\}$ and $I_2 = \{2, 3\}$ define the function: $y = \frac{1}{T}(x_1 \bar x_2 + \bar x_1 x_2)$. The user set parameter $T$ decides the resolution of the regression function. Notice that each product in the summation either evaluates to $0$ or $1$. This means that a larger $T$ requires more literal products to reach a particular value $y$. Thus, increasing $T$ makes the regression function increasingly fine-grained. In the following, we will formulate and refer to the products as \emph{conjunctive clauses}, as is typical for the regular TM. The value $c_j$ of each product is then a conjunction of literals:
\begin{equation}\label{Eq1}
c_j = \prod_{k \in I_j} l_k = \bigwedge_{k \in I_j} l_k.
\end{equation}
Finally, note that the number of conjunctive clauses $m$ in the regression function also is a user set parameter, which decides the expression power of the RTM.

{\bf Tsetlin Automata Teams:} The composition of each clause is decided by a team of Tsetlin Automata (TAs) \cite{Tsetlin1961}. Each TA is a finite state automaton that has $2N$ states. The state decides which action the TA performs, and it is updated from feedback using a linear strategy. The aim of a TA is to find the optimal action as quickly as possible, trading off exploration against exploitation. There are $2 \times o$ number of TAs per clause $j$. Each of these TAs is associated with a particular literal $k$ and decides whether to \textit{include} or \textit{exclude} that literal in the clause. The decision depends on the current state of the TA, denoted $a_{j,k} \in \{1, \ldots, 2N\}$. States from 1 to $N$ produce an \textit{exclude} action and states from $N+1$ to $2N$ produce an \textit{include} action. Accordingly, the set of indexes $I_j$ can be defined as $I_j = \left\{ k | a_{j,k} > N,  1 \le k \le 2o\right\}$. The states of all of the TAs are organized as an $m \times 2o$ matrix $\mathbf{A}$: $\mathbf{A} = (a_{j,k}) \in \{1, \ldots, 2N\}^{m \times 2o}$ where $m$ is the number of clauses. 

{\bf Overall Learning Procedure:} In the training phase, the TAs learn to decide between \textit{include} and \textit{exclude} actions. This is done through an online reinforcement scheme that updates the state matrix $\mathbf{A}$ by processing one training example $(\hat{X}_i, \hat{y}_i)$ at a time, drawn from a set of training examples. The scheme coordinates the TA team as a whole, since all of the TAs in all of the clauses jointly contribute to produce the final output $y$, for every training example. 

To this end, the RTM employs two kinds of feedback, Type~I and Type~II, further defined below. Type I feedback triggers TA state changes that eventually make a clause output $1$ for the given training example $\hat{X}_i$. Conversely, Type II feedback triggers state changes that eventually make the clause output $0$. Thus, overall, regression error can be systematically reduced by carefully distributing Type I and Type II feedback:
        \begin{equation}\label{eq9}
        Feedback = 
        \begin{cases}
        \text{Type I},   \;\;\;\;\;\;\;\;   & \text{if \;\; } y  <  \hat{y}_i ,\\
        \text{Type II},      \;\;\;\;\;\; \;\;   & \text{if \;\; } y > \hat{y}_i .
        \end{cases}
        \end{equation}
 In effect, the number of clauses that evaluates to 1 is increased when the predicted output is less than the target output ($y < \hat{y}_i$) by providing Type I feedback to the clauses.  Conversely, Type II feedback is applied to decrease the number of clauses that evaluates to 1 when the predicted output is higher than the target output ($y > \hat{y}_i$). Since the TAs learn conservatively through state changes, the above procedure gradually reduces regression error, in small steps. 
 
 {\bf Activation Probability:} Feedback is handed out stochastically to regulate learning. If the regression error is large, the RTM compensates by giving feedback to more clauses. Specifically, the probability of giving a clause $j$ feedback is proportionally to the absolute error of the prediction. Below, the variable $p_j$ decides whether a particular clause $j$ is given feedback:
        \begin{equation}\label{eq10}
        p_{j} =
        \begin{cases}
        1 & \text{with probability } \frac{\mid y - \hat{y}_i \mid}{T}, \\
        0 & \text{otherwise}.
        \end{cases}
        \end{equation}
As seen, in addition to the absolute regression error, the user set resolution $T$ also decides the frequency of the feedback. A higher $T$ reduces the overall probability of feedback, resulting in more conservative learning. Which clauses are activated for feedback is stored in the matrix $\mathbf{P} = (p_j) \in \{0,1\}^m$.


{\bf Type I feedback:} Type I feedback subdivides into Type Ia and Type Ib. Type~Ia reinforces \textit{include} actions of TAs whose corresponding literal value is 1, however, only when the clause output is $1$. This makes the clause gradually resemble the input $\hat{X}_i$ itself. The purpose is to capture the underlying frequent patterns governing the regression.  Type~Ib combats over-fitting by reinforcing \textit{exclude} actions of TAs when the corresponding literal is $0$ or when the clause output is $0$.

Type Ib feedback is provided to TAs stochastically using a user set parameter $s$ (s $\geq 1$). That is, the decision whether the $k^{th}$ TA of the $j^{th}$ clause receives Type Ib feedback ($q_{j,k}$) is stochastically made as follows,

\begin{equation}
q_{j,k} =
\begin{cases}
1 & \text{with probability } \frac{1}{s},\\
0 & \text{otherwise}.
\end{cases}
\end{equation}

Using the complete set of conditions, the TAs selected for Type Ia feedback are singled out by the indexes ${I}^{\text{Ia}} = \left\{(j,k) | l_{k} = 1 \land c_j = 1 \land p_j = 1  \right\}.$ Similarly, TAs selected for Type Ib are ${I}^{\text{Ib}} = \left\{(j,k) | (l_{k} = 0 \lor c_j = 0) \land p_{j} = 1 \land q_{j,k} = 1 \right\}.$ 

Once the TAs have been targeted for Type Ia and Type Ib feedback, their states are updated. Available updating operators are $\oplus$ and $\ominus$, where $\oplus$ adds 1 to the current state while $\ominus$ subtracts 1. Thus, before a new learning iterations starts, the states in the matrix $\mathbf{A}$ are updated as follows: $\mathbf{A} \leftarrow \left( \mathbf{A} \oplus {{I}^{\text{Ia}}}\right) \ominus {{I}^{\text{Ib}}}$.

{\bf Type II feedback:} Type II feedback eventually changes the output of a clause from $1$ to $0$, for a specific input $\hat{X}_i$. The goal is to increase the discrimination power of the clause. This is achieved simply  by including one or more of the literals that take the value $0$ for $\hat{X}_i$. The indexes of TAs selected for Type II can thus be singled out as ${I}^{\text{II}} = \left\{(j,k) | l_{k} = 0 \land c_j = 1 \land p_{j} = 1 \right\}$. Accordingly, the states of the TAs are updated as follows: $\mathbf{A} \leftarrow \mathbf{A}\oplus {I}^{\text{II}}$. By increasing the TA states, eventually, one or more TAs switch from excluding their literals to including them, rendering the clause output $0$.
    
\section{Stochastic Searching on the Line}
Stochastic searching on the line, also referred to as stochastic point location (SPL) was pioneered by Oommen in 1997 \cite{oommen1997stochastic}. SPL is a fundamental optimization problem where one tries to locate an unknown unique point within a given interval. The only available information for the Learning Mechanism (LM) is the possibly faulty feedback provided by the attached environment ($E$). According to the feedback, LM moves right or left from its current location in a discretized solution space.

The task at hand is to determine the optimal value $\lambda^*$ of a variable $\lambda$, assuming that the environment is informative. That is, that it provides the correct direction of $\lambda^*$ with probability $p > 0.5$. Here, the value of $p$ reflects the "effectiveness" of the environment. In SPL, $\lambda$ is assume to be any number in the interval $[0, 1]$. The SPL scheme of Oommen discretizes the solution space by subdividing the unit interval into $N$ steps, \{$0, 1/N, 2/N, ..., (N-1)/N, 1$\}. Hence, $N$ defines the resolution of the learning scheme.

The current guess, $\lambda(n)$, is updated according to the feedback from the environment as follows: 

\begin{equation}\label{eq11}
\lambda(n+1) = 
\begin{cases}
\lambda(n) + 1/N,   \;\;\;\;\;\;\;\;   & \text{if \;\; } E(n)=1 \text{ and } 0\leqslant \lambda(n)<1 \; ,\\
\lambda(n) - 1/N,      \;\;\;\;\;\; \;\;   & \text{if \;\; } E(n)=0 \text{ and } 0< \lambda(n)\leqslant1 \;.
\end{cases}
\end{equation}

\begin{equation}\label{eq12}
\lambda(n+1) = 
\begin{cases}
\lambda(n),   \;\;\;\;\;\;\;\;   & \text{if \;\; } \lambda(n)=1 \text{ and } E(n)=1  \; ,\\
\lambda(n),      \;\;\;\;\;\; \;\;   & \text{if \;\; } \lambda(n)=0 \text{ and } E(n)=0 \;.
\end{cases}
\end{equation}

Here, $\lambda(n)$ is the value of $\lambda$ at time step $n$. The feedback from the environment has been binarized, where $E(n) = 1$ is the environment suggestion to increase the value of $\lambda$ and $E(n) = 0$ is the environment suggestion to decrease the value of $\lambda$. Asymptotically, the learning mechanics is able to find a value arbitrarily close to $\lambda^*$ when $N \rightarrow \infinity$ and $n \rightarrow \infinity$.

\section{Regression Tsetlin Machine with Weighted Clauses}

We now introduce clauses with integer weights to provide a more compact representation of the regression function. In contrast to the weighting scheme proposed by Phoulady et al. for the standard TM \cite{phoulady2020weighted}, we represent the weights as integers, leveraging stochastic searching one the line. The purpose is to eliminate multiplication from the weight updating, relying purely on increment and decrement operations. In addition to the computational benefits this entails, we also postulate that integer weighted clauses are more interpretable than real-valued ones because they can be seen as multiple copies of the same clause. 

\begin{figure}[t]
\centering
\includegraphics[width=8cm]{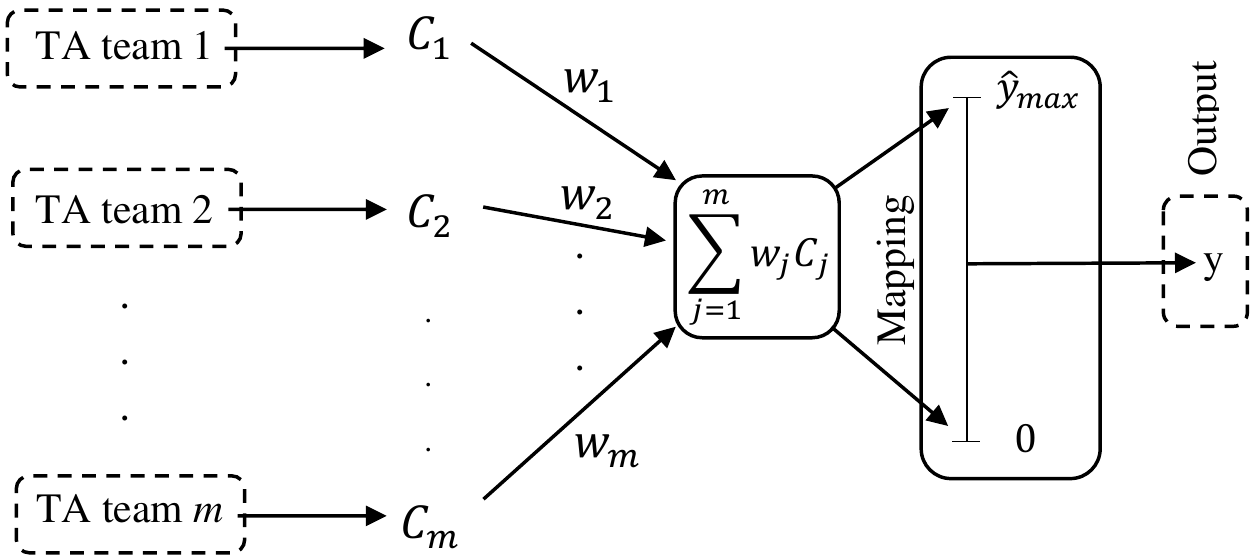}
\caption{The RTM with integer weights.} \label{WRTM}
\end{figure}

{\bf Regression function:} The regression function for the integer weighted RTM attaches a weight $w_j$ to each clause output $c_j$, $j = 1, ...,m$. Consequently, the regression output can be computed according to Eq. \ref{eq13} (as illustrated in Fig. \ref{WRTM}):

\begin{equation}\label{eq13}
y = \frac{1}{T} \sum_{j=1}^m  w_j \prod_{k \in I_j} l_k. 
\end{equation}

{\bf Weight learning:} Our approach to learning the weight of each clause is similar to SPL. However, the solution space of each weight is $[0, \infinity]$, while the resolution of the learning scheme is $N = 1$. The weight attached to a clause is updated when the clause receives Type Ia feedback or Type II feedback. The weight updating procedure is summarized in Algorithm 1. Here, $w_j(n)$ is the weight of clause $j$ at the $n^{th}$ training round.

\begin{algorithmic}
\vspace{3mm}

\STATE \textbf{Algorithm 1: Round $n$ updating of clause weights}\\
\STATE \textbf{Initialization (round $0$):} $w_j(0) \leftarrow 0, j = 1, \ldots, m$\\
\STATE \textbf{Initialization (round $n$):} $y$ is calculated according to Eq. \ref{eq13}.
\FOR{$j = 1, ...,m$}
\IF {$y(n) < \hat{y}_i(n) \land c_j(n) = 1 \land p_j(n) = 1$} 
    \STATE $w_j(n+1) \leftarrow w_j(n) + N$
\ELSIF{$y(n) > \hat{y}_i(n) \land p_j(n) = 1 \land w_j(n) > 0$}
    \STATE $w_j(n+1) \leftarrow w_j(n) - N$
\ELSE
    \STATE $w_j(n+1) \leftarrow w_j(n)$
\ENDIF
\ENDFOR
\STATE \textbf{Return} $w_j(n+1), j = 1, \ldots, m$
\vspace{3mm}
\end{algorithmic}
Note that since weights in this study can take any value higher than or equal to 0, an unwanted clause can be turned off by setting its weight to 0. Further, sub-patterns that have a large impact on the calculation of $y$ can be represented with a correspondingly larger weight. 

\section{Empirical Evaluation}
In this section, we study the behavior of the RTM with integer weighting (RTM-IW) using six artificial datasets similar to the datasets presented in \cite{abeyrathnartm}, in comparison with a standard RTM and a real-value weighted RTM (RTM-RW). We use Mean Absolute Error ($MAE$) to measure performance.

\subsection{Artificial Datasets}
Dataset I contains 2-bit feature input. The output, in turn, is 100 times larger than the decimal
value of the binary input (e.g., the input [1, 0] produces the output $200$).
The training set consists of $8000$ samples while the testing set consists of $2000$
samples, both without noise. Dataset II contains the same data as Dataset I,
except that the output of the training data is perturbed to introduce noise. For
Dataset III we introduce 3-bit input, without noise, and for Dataset IV we have
3-bit input with noisy output. Finally, Dataset V has 4-bit input without noise,
and Dataset VI has 4-bit input with noise. Each input feature have been generated independently with equal probability of taking either the value $0$ or $1$, producing a uniform distribution of bit values.

\subsection{Results and Discussion}

\begin{figure}[!t]
\centering
\includegraphics[width=5cm]{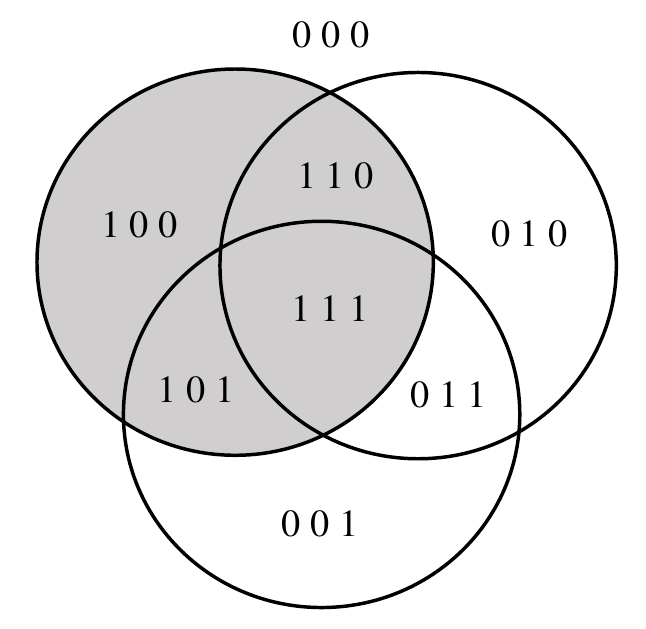}
\caption{Pattern distribution of the 3-bits input dataset.} \label{figpatterns}
\end{figure}

The pattern distribution of the artificial data was analyzed in the original RTM study using Fig. \ref{figpatterns}, which illustrates the pattern distribution for the case of 3-bit input. As depicted, there are eight unique sub-patterns. The RTM is able to capture the complete set of sub-patterns utilizing no more than three types of clauses, i.e., (1 \ding{83} \ding{83}), (\ding{83} 1 \ding{83}), (\ding{83} \ding{83} 1)\footnote{Here, \ding{83} means an input feature that can take an arbitrary value, either 0 or 1.}. However, to produce the correct output, as found in the training and testing data, each clause must be duplicated multiple times, depending on the input pattern. For instance, Dataset III requires \emph{seven} clauses to represent the three different patterns it contains, namely, (4 × (1 \ding{83} \ding{83}), 2 × (\ding{83} 1 \ding{83}), 1 × (\ding{83} \ding{83} 1))\footnote{In this expression, \enquote{\emph{four} clauses to represent the pattern (1 \ding{83} \ding{83})} is written as \enquote{4 × (1 \ding{83} \ding{83})}}. So, with e.g. the input [1, 0, 1], four clauses which represent the pattern (1 \ding{83} \ding{83}) and one clause which represents the pattern (\ding{83} \ding{83} 1) activate to correctly output $500$ (after normalization).

Notably, it turns out that the RTM-IW requires even fewer clauses to capture the sub-patterns in the above data, as outlined in Table \ref{tab1}. Instead of having multiple clauses to represent one sub-pattern, RTM-IW utilizes merely one clause with the correct weight to do the same job. The advantage of the proposed integer weighting scheme is thus apparent. It learns the correct weight of each clause, so that it achieves an MAE of zero. Further, it is possible to ignore redundant clauses simply by giving them the weight zero. For the present dataset, for instance, increasing $m$ while keeping the same resolution, $T=7$, does not impede accuracy. The RTM-RW on the other hand struggles to find the correct weights, and fails to minimize MAE. Here, the real valued weights were updated with a learning rate of $\alpha = 0.01$, determined using a binary hyper-parameter search.

\begin{table}[!t]
\centering
\caption{Behavior comparison of different RTM schemes on Dataset III.}\label{tab1}
\begin{tabular}{c|c|c|c|c|c|c|c|c|c|}
\cline{2-10}
                                                                                                         & $m$                  & $T$                  & Pattern & \begin{tabular}[c]{@{}c@{}} ${I}_j$\end{tabular} & \begin{tabular}[c]{@{}c@{}} $\bar{I}_j$\end{tabular} & \begin{tabular}[c]{@{}c@{}}No. of Clauses\\ Required\end{tabular} & $w_j$ & \begin{tabular}[c]{@{}c@{}}Training\\ MAE\end{tabular} & \begin{tabular}[c]{@{}c@{}}Testing \\ MAE\end{tabular} \\ \hline
\multicolumn{1}{|c|}{\multirow{3}{*}{RTM}}                                                               & \multirow{3}{*}{7} & \multirow{3}{*}{7} & (1 \ding{83} \ding{83})   & {\{}1{\}}                                             & {\{} {\}}                                             & 4                                                                 & -      & \multirow{3}{*}{0}                                     & \multirow{3}{*}{0}                                     \\ \cline{4-8}
\multicolumn{1}{|c|}{}                                                                                   &                    &                    & (\ding{83} 1 \ding{83})    & {\{}2{\}}                                             & {\{} {\}}                                             & 2                                                                 & -      &                                                        &                                                        \\ \cline{4-8}
\multicolumn{1}{|c|}{}                                                                                   &                    &                    & (\ding{83} \ding{83} 1)    & {\{}3{\}}                                             & {\{} {\}}                                             & 1                                                                 & -      &                                                        &                                                        \\ \hline
\multicolumn{1}{|c|}{\multirow{3}{*}{\begin{tabular}[c]{@{}c@{}}RTM-IW\end{tabular}}} & \multirow{3}{*}{3} & \multirow{3}{*}{7} & (1 \ding{83} \ding{83})   & {\{}1{\}}                                             & {\{} {\}}                                             & 1                                                                 & 4      & \multirow{3}{*}{0}                                     & \multirow{3}{*}{0}                                     \\ \cline{4-8}
\multicolumn{1}{|c|}{}                                                                                   &                    &                    & (\ding{83} 1 \ding{83})    & {\{}2{\}}                                             & {\{} {\}}                                             & 1                                                                 & 2      &                                                        &                                                        \\ \cline{4-8}
\multicolumn{1}{|c|}{}                                                                                   &                    &                    & (\ding{83} \ding{83} 1)    & {\{}3{\}}                                             & {\{} {\}}                                             & 1                                                                 & 1      &                                                        &                                                        \\ \hline
\multicolumn{1}{|c|}{\multirow{3}{*}{\begin{tabular}[c]{@{}c@{}}RTM-RW\end{tabular}}}      & \multirow{3}{*}{3} & \multirow{3}{*}{7} & (1 \ding{83} \ding{83})   & {\{}1{\}}                                             & {\{} {\}}                                             & 1                                                                 & 3.987  & \multirow{3}{*}{1.857}                                 & \multirow{3}{*}{1.799}                                 \\ \cline{4-8}
\multicolumn{1}{|c|}{}                                                                                   &                    &                    & (\ding{83} 1 \ding{83})    & {\{}2{\}}                                             & {\{} {\}}                                             & 1                                                                 & 2.027  &                                                        &                                                        \\ \cline{4-8}
\multicolumn{1}{|c|}{}                                                                                   &                    &                    & (\ding{83} \ding{83} 1)    & {\{}3{\}}                                             & {\{} {\}}                                             & 1                                                                 & 0.971  &                                                        &                                                        \\ \hline
\end{tabular}
\end{table}

\begin{figure}[t]
\centering
\includegraphics[width=9cm]{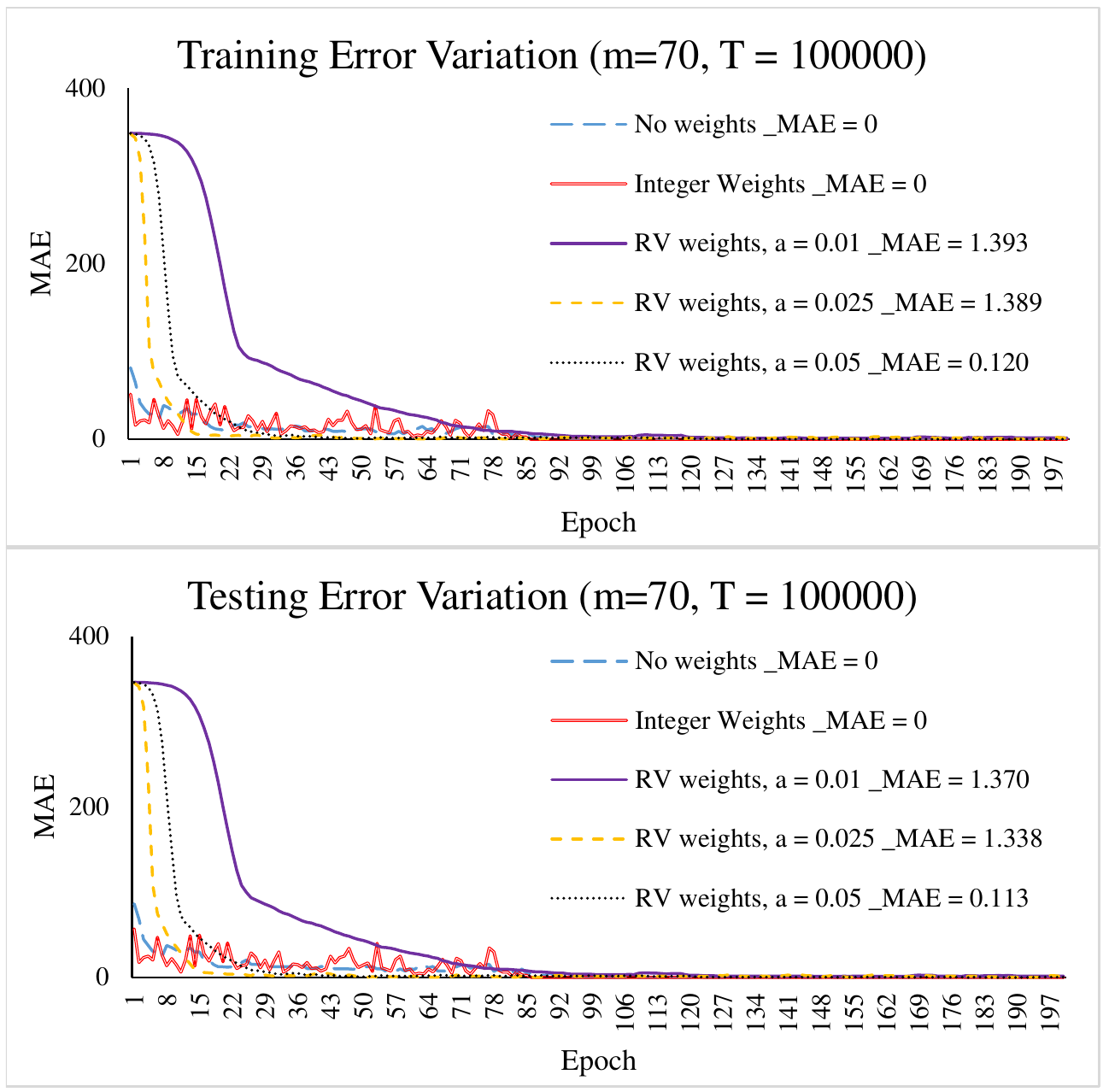}
\caption{The training and testing error variation per training epoch for different RTM schemes.} \label{figplot1}
\end{figure}

Fig. \ref{figplot1} casts further light on learning behaviour by reporting training and testing error per epoch for the three different RTM schemes for Dataset III, with $m=70$ and $T=100000$. As seen, both RTM and RTM-IW obtain relatively low MAE after just one training epoch, eventually reaching MAE zero (training and testing MAE at end of training are given in the legend of each graph). RTM-RW, on the other hand, starts off with a much higher MAE, which is drastically decreasing over a few epochs, however, fails to reach MAE $0$ after becoming asymptotically stable.

We also studied the effect of $T$ on performance with noise free data by varying $T$, while fixing the number of clauses $m$. For instance, RTM was able to reach a training MAE of $1.9$ and a testing error of $2.1$ with $m = T = 300$ on Dataset III  \cite{abeyrathnartm}. For the same dataset, RTM-IW can reach a training error of $0.19$ and a testing error of $1.87$ with $m = 200$ and $T = 2000$. Further, for  $m = 200$ and $T = 20~000$, training error drops to $0.027$ and testing error drops to $0.027$. Finally, by increasing $T$ to $200~000$ training error falls to $0.0003$ while testing error stabilises at $0.0002$.

To further compare the performance of RTM-IW with RTM and RTM-RW, each approach was further evaluated using a wide rage of $m$ and $T$ settings. Representative training and testing MAE for all datasets are summarized in Table \ref{tab2}. Here, the number of clauses used with each dataset is also given. The $T$ for the original RTM is equal to the number of clauses, while for the RTM with weights $T$ is simply $100$ times that number. 

\begin{table}[!t]
\centering
\caption{Training and testing MAE after 200 training epochs by various methods with different $m$ and $T$.}\label{tab2}
\begin{tabular}{|c|c|c|c|c|c|c|c|}
\hline
\multirow{2}{*}{Dataset} & \multirow{2}{*}{m} & \multicolumn{3}{c|}{Training MAE}                                                                                                                                              & \multicolumn{3}{c|}{Testing MAE}                                                                                                                                               \\ \cline{3-8} 
                         &                          & \begin{tabular}[c]{@{}c@{}}RTM\end{tabular} & \begin{tabular}[c]{@{}c@{}}RTM-RW\end{tabular} & \begin{tabular}[c]{@{}c@{}}RTM-IW\end{tabular} & \begin{tabular}[c]{@{}c@{}}RTM\end{tabular} & \begin{tabular}[c]{@{}c@{}}RTM-RW\end{tabular} & \begin{tabular}[c]{@{}c@{}}RTM-IW\end{tabular} \\ \hline
\multirow{7}{*}{I}       & 3                        & 0.0000                                                & 0.5898                                                & 0.0000                                                     & 0.0000                                                & 0.5815                                                & 0.0000                                                     \\ \cline{2-8} 
                         & 10                       & 7.8000                                                & 0.1650                                                & 0.1655                                                     & 7.6000                                                & 0.1659                                                & 0.1653                                                     \\ \cline{2-8} 
                         & 30                       & 0.0000                                                & 0.0378                                                & 0.0000                                                     & 0.0000                                                & 0.0378                                                & 0.0000                                                     \\ \cline{2-8} 
                         & 100                      & 0.8000                                                & 0.0040                                                & 0.0149                                                     & 0.8000                                                & 0.0039                                                & 0.0151                                                     \\ \cline{2-8} 
                         & 500                      & 0.5000                                                & 0.0013                                                & 0.0017                                                     & 0.5000                                                & 0.0013                                                & 0.0017                                                     \\ \cline{2-8} 
                         & 1000                     & 0.2000                                                & 0.0005                                                & 0.0008                                                     & 0.2000                                                & 0.0005                                                & 0.0008                                                     \\ \cline{2-8} 
                         & 4000                     & 0.3000                                                & 0.0002                                                & 0.0002                                                     & 0.3000                                                & 0.0002                                                & 0.0002                                                     \\ \hline
\multirow{7}{*}{II}       & 3                        & 7.2000                                                & 7.4157                                                & 7.2630                                                     & 5.0000                                                & 5.6083                                                & 5.2979                                                     \\ \cline{2-8} 
                         & 10                       & 11.0000                                               & 7.7618                                                & 6.8047                                                     & 10.6000                                               & 6.4026                                                & 4.8627                                                     \\ \cline{2-8} 
                         & 30                       & 8.8000                                                & 6.1403                                                & 7.2517                                                     & 7.1000                                                & 3.7591                                                & 5.2997                                                     \\ \cline{2-8} 
                         & 100                      & 5.4000                                                & 5.8588                                                & 5.9486                                                     & 1.2000                                                & 2.9511                                                & 2.9288                                                     \\ \cline{2-8} 
                         & 500                      & 5.5000                                                & 5.6255                                                & 5.6483                                                     & 2.7000                                                & 2.3199                                                & 2.3893                                                     \\ \cline{2-8} 
                         & 1000                     & 5.2000                                                & 5.7425                                                & 5.5383                                                     & 1.6000                                                & 2.4535                                                & 2.0222                                                     \\ \cline{2-8} 
                         & 4000                     & 5.4000                                                & 5.6552                                                & 5.5673                                                     & 1.8000                                                & 2.3977                                                & 2.1777                                                     \\ \hline
\multirow{7}{*}{III}       & 7                        & 0.0000                                                & 2.2296                                                & 1.1723                                                     & 0.0000                                                & 2.2173                                                & 1.1710                                                     \\ \cline{2-8} 
                         & 20                       & 14.6000                                               & 1.0232                                                & 0.4873                                                     & 14.2000                                               & 1.0362                                                & 0.4933                                                     \\ \cline{2-8} 
                         & 70                       & 0.0000                                                & 0.2920                                                & 0.1889                                                     & 0.0000                                                & 0.2946                                                & 0.1894                                                     \\ \cline{2-8} 
                         & 300                      & 1.9000                                                & 0.1037                                                & 0.0776                                                     & 2.1000                                                & 0.1057                                                & 0.0776                                                     \\ \cline{2-8} 
                         & 700                      & 1.0000                                                & 0.0130                                                & 0.0435                                                     & 1.0000                                                & 0.0131                                                & 0.0438                                                     \\ \cline{2-8} 
                         & 2000                     & 1.0000                                                & 0.0117                                                & 0.0034                                                     & 1.2000                                                & 0.0118                                                & 0.0034                                                     \\ \cline{2-8} 
                         & 5000                     & 0.9000                                                & 0.0097                                                & 0.0014                                                     & 1.0000                                                & 0.0100                                                & 0.0014                                                     \\ \hline
\multirow{7}{*}{IV}       & 7                        & 7.4000                                                & 7.7023                                                & 8.0185                                                     & 5.0000                                                & 5.9550                                                & 6.2355                                                     \\ \cline{2-8} 
                         & 20                       & 13.8000                                               & 7.8625                                                & 9.8444                                                     & 14.5000                                               & 6.0067                                                & 8.4991                                                     \\ \cline{2-8} 
                         & 70                       & 6.6000                                                & 7.3648                                                & 7.6019                                                     & 4.2000                                                & 5.7352                                                & 5.5316                                                     \\ \cline{2-8} 
                         & 300                      & 5.8000                                                & 5.7999                                                & 5.6845                                                     & 3.3000                                                & 2.2255                                                & 2.2342                                                     \\ \cline{2-8} 
                         & 700                      & 5.9000                                                & 5.5514                                                & 5.5324                                                     & 3.4000                                                & 1.9676                                                & 2.1493                                                     \\ \cline{2-8} 
                         & 2000                     & 5.6000                                                & 5.7311                                                & 5.3726                                                     & 1.9000                                                & 2.5195                                                & 1.2801                                                     \\ \cline{2-8} 
                         & 5000                     & 5.5000                                                & 5.6350                                                & 5.4119                                                     & 2.7000                                                & 2.2517                                                & 1.5015                                                     \\ \hline
\multirow{7}{*}{V}       & 7                        & 9.8000                                                & 77.9091                                               & 64.8378                                                    & 9.9000                                                & 79.3980                                               & 58.3262                                                    \\ \cline{2-8} 
                         & 15                       & 0.0000                                                & 2.3127                                                & 1.5787                                                     & 0.0000                                                & 2.3178                                                & 1.5575                                                     \\ \cline{2-8} 
                         & 70                       & 1.7000                                                & 0.7583                                                & 0.7583                                                     & 1.8000                                                & 0.7527                                                & 0.7527                                                     \\ \cline{2-8} 
                         & 150                      & 0.0000                                                & 0.2649                                                & 0.1233                                                     & 0.0000                                                & 0.2657                                                & 0.1242                                                     \\ \cline{2-8} 
                         & 700                      & 0.2000                                                & 0.0441                                                & 0.0315                                                     & 0.3000                                                & 0.0436                                                & 0.0313                                                     \\ \cline{2-8} 
                         & 1500                     & 0.2000                                                & 0.0373                                                & 0.0200                                                     & 0.2000                                                & 0.0378                                                & 0.0200                                                     \\ \cline{2-8} 
                         & 4000                     & 0.2000                                                & 0.0174                                                & 0.0051                                                     & 0.2000                                                & 0.0174                                                & 0.0050                                                     \\ \hline
\multirow{7}{*}{VI}       & 7                        & 79.8000                                               & 58.1584                                               & 51.8698                                                    & 78.0000                                               & 58.4676                                               & 53.1777                                                    \\ \cline{2-8} 
                         & 15                       & 51.4000                                               & 11.2369                                               & 11.8776                                                    & 50.1000                                               & 9.6501                                                & 10.7141                                                    \\ \cline{2-8} 
                         & 70                       & 13.1000                                               & 8.0054                                                & 6.6716                                                     & 12.5000                                               & 6.2236                                                & 4.5814                                                     \\ \cline{2-8} 
                         & 150                      & 10.3000                                               & 6.5524                                                & 7.2056                                                     & 8.5000                                                & 4.2723                                                & 5.2055                                                     \\ \cline{2-8} 
                         & 700                      & 5.5000                                                & 6.1536                                                & 5.8699                                                     & 3.5000                                                & 3.5392                                                & 3.1662                                                     \\ \cline{2-8} 
                         & 1500                     & 5.3000                                                & 5.9487                                                & 5.5769                                                     & 2.7000                                                & 3.1904                                                & 2.2793                                                     \\ \cline{2-8} 
                         & 4000                     & 5.4000                                                & 5.5568                                                & 5.4816                                                     & 2.8000                                                & 2.3064                                                & 2.1397                                                     \\ \hline
\end{tabular}
\end{table}

As seen, the training and testing MAE reach zero when the RTM operates with noise free data. Similar performance can be seen with RTM-IW for Dataset I, but not for the other two noise free datasets. However, as seen, MAE approaches zero with increasing number of clauses $m$.

For noisy data (Dataset IV and Dataset VI), the minimum training MAE achieved by RTM is $5.500$, obtained with $m=5000$ clauses. The RTM-IW, on the other hand,  obtains a lower MAE of $5.3726$ with less than half of the clauses ($m = 2000$). Similarly, on Dataset VI, RTM-IW outperforms the lowest RTM MAE of $2.2793$, both using $1500$ clauses.

The accuracy of RTM-IW in comparison with RTM-RW is less clear, with quite similar MAE for all of the datasets. The average testing MAE across all of the datasets, however, reveals that the average MAE of RTM-IW is lower than that of the RTM-RW ($4.682$ vs $5.417$).

\begin{figure}[!t]
\begin{subfigure}{.5\textwidth}
  \centering
  \includegraphics[width=1\linewidth]{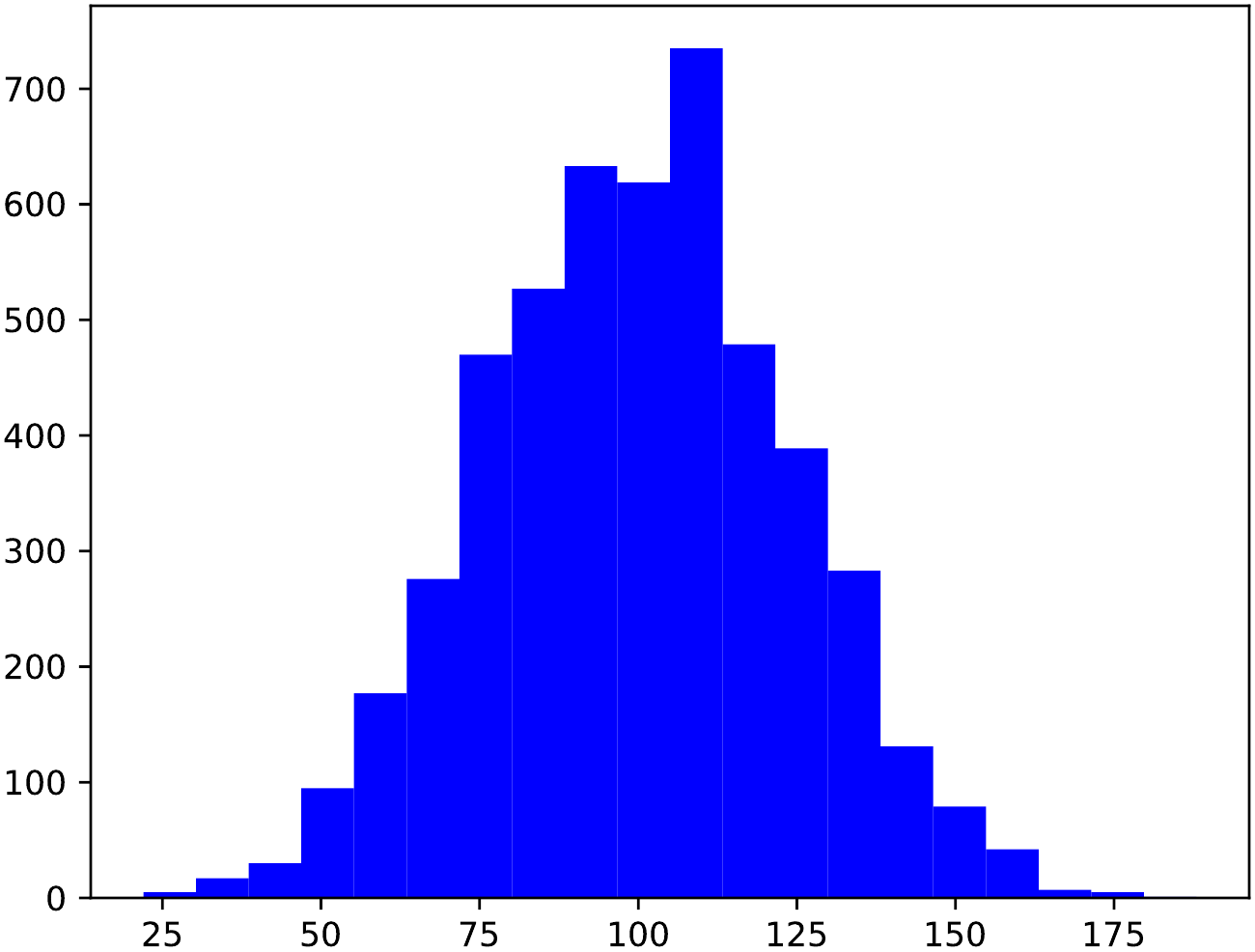}  
  \caption{Weights for Dataset III with RTM-IW}
  \label{fig:sub-first}
\end{subfigure}
\begin{subfigure}{.5\textwidth}
  \centering
  \includegraphics[width=1\linewidth]{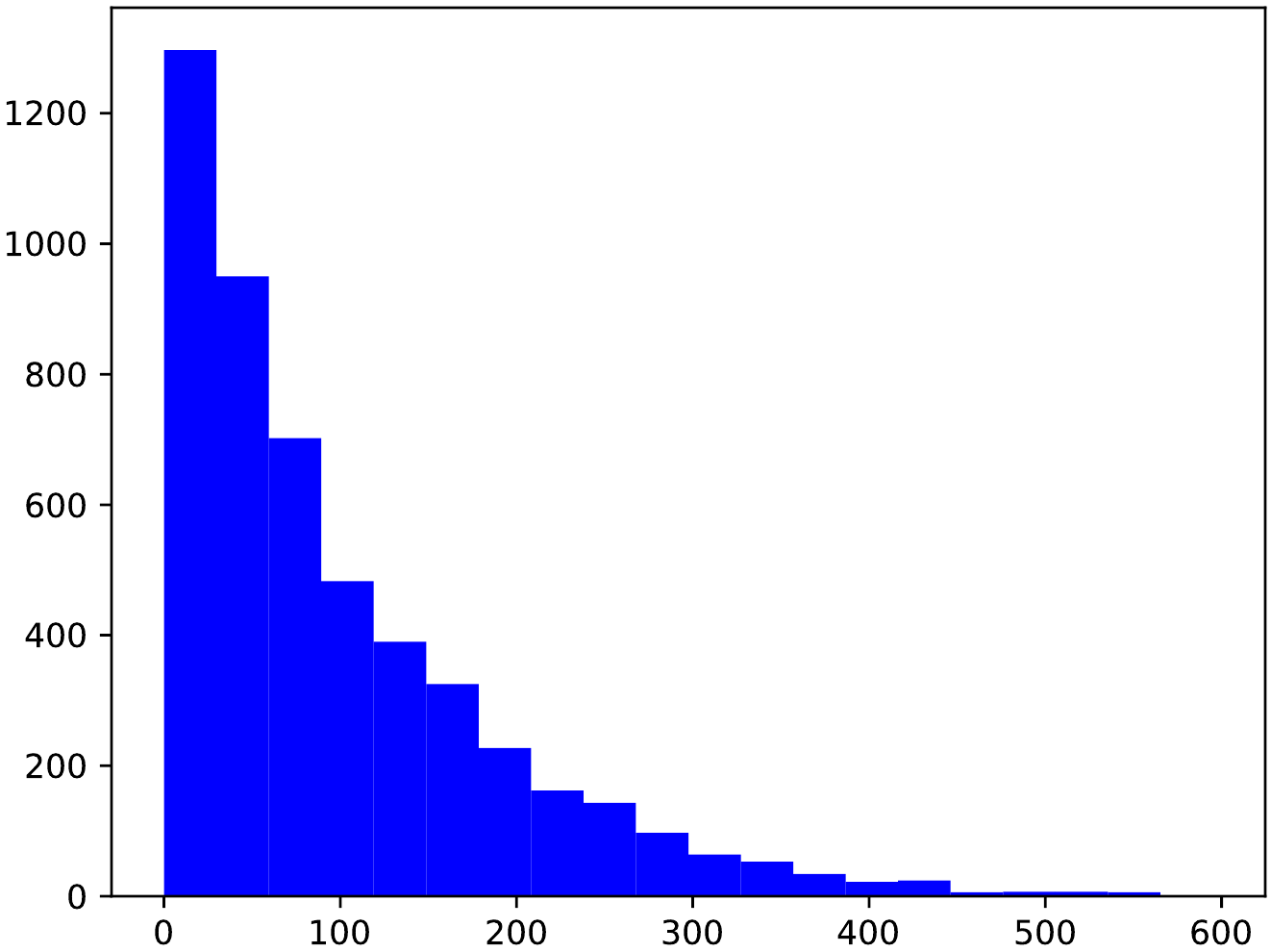}  
  \caption{Weights for Dataset IV with RTM-IW}
  \label{fig:sub-third}
\end{subfigure}
\caption{The distribution of weights learnt for Dataset III and IV by RTM-IW with $m=5000$ and $T=500000$}
\label{figweights}
\end{figure}

Finally, Fig. \ref{figweights} shows the distribution of weights at end of training when RTM-IW utilize the highest number of clauses from Table \ref{tab2} for  Dataset III and Dataset IV. As seen, weights for the Dataset III have a normally distributed shape, with the mean found around $100$. Surprisingly, almost zero clauses have been turned off by setting their weights to zero while working with noise free data. The weight distribution for Dataset IV, on the other hand, shows that larger potion of clauses have been turned off while working with noisy data. Further, surprisingly, the distribution now has an exponential form. The weight distributions for the other datasets behave similarly.

\section{Conclusion}
In this paper, we presented a new weighting scheme for the Regression Tsetlin Machine (RTM), RTM with Integer Weights (RTM-IW). The weights attached to the clauses helps the RTM represent sub-patterns in a more compact way. Since the weights are integer, interpretability is improved through a more compact representation of the clause set. We also presented a new weight learning scheme based on stochastic searching on the line, integrated with the Type I and Type II feedback of the RTM. The RTM-IW obtains on par or better accuracy with fewer number of clauses compared to RTM without weights. It also performs competitively in comparison with an alternative RTM with real-valued weights.

\bibliographystyle{ieeetr}
\begin{bibliography}{sample}
\end{bibliography}

\end{document}